\pdfoutput=1

\documentclass[11pt]{article}

\usepackage{ACL2023}

\usepackage{times}
\usepackage{latexsym}

\usepackage[T1]{fontenc}

\usepackage[utf8]{inputenc}

\usepackage{microtype}

\usepackage{inconsolata}

\usepackage{algorithm}
\usepackage{algpseudocode}
\usepackage{amsmath}
\usepackage{graphicx}
\usepackage{soul}
\usepackage{colortbl}
\usepackage{multicol}
\usepackage{booktabs}
\usepackage{multirow}
\usepackage{arydshln}
\usepackage[capitalize,noabbrev]{cleveref}

\algrenewcommand\algorithmicindent{1.15em}
\algnewcommand{\LeftComment}[1]{\\ \hspace{\algorithmicindent}\hspace{\algorithmicindent}\(\triangleright\) #1}
\newcolumntype{C}[1]{>{\centering\arraybackslash}m{#1}}
\newcolumntype{L}[1]{>{\raggedright\arraybackslash}m{#1}}

\definecolor{lred}{rgb}{0.925, 0.816, 0.863}
\definecolor{lgreen}{rgb}{0.847,0.922,0.831}
\definecolor{lyellow}{rgb}{1.0,0.882,0.584}
\definecolor{lblue}{rgb}{0.761,0.835,0.961}

\newcommand{\remove}{\protect\sethlcolor{lred}\protect\hl{\textit{remove}}}
\newcommand{\present}{\protect\sethlcolor{lgreen}\protect\hl{\textit{change tense to present simple}}}
\newcommand{\swap}{\protect\sethlcolor{lyellow}\protect\hl{\textit{swap with synonym}}}
\newcommand{\add}{\protect\sethlcolor{lblue}\protect\hl{\textit{add synonym}}}

\newcommand{\cremove}[1]{\protect\sethlcolor{lred}\protect\hl{#1}}
\newcommand{\cpresent}[1]{\protect\sethlcolor{lgreen}\protect\hl{#1}}
\newcommand{\cswap}[1]{\protect\sethlcolor{lyellow}\protect\hl{#1}}
\newcommand{\cadd}[1]{\protect\sethlcolor{lblue}\protect\hl{#1}}

\title{Query Rewriting for Effective Misinformation Discovery}

\author{
    Ashkan Kazemi\textsuperscript{*1}, Artem Abzaliev\textsuperscript{*1}, Naihao Deng\textsuperscript{1}\\{\bf Rui Hou\textsuperscript{2}, Scott A. Hale\textsuperscript{3,4}, Verónica Pérez-Rosas\textsuperscript{1}, Rada Mihalcea\textsuperscript{1}}\\
    \textsuperscript{1}University of Michigan, Meta AI\textsuperscript{2}, Meedan\textsuperscript{3}, Oxford Internet Institute\textsuperscript{4}\\
    \texttt{\{ashkank, abzaliev, dnaihao\}@umich.edu}\\ \texttt{rayhou@meta.com, scott@meedan.com, \{vrncapr, mihalcea\}@umich.edu}\\
    \textsuperscript{*}\textit{contributed equally}\\
}

\begin{document}
\maketitle
\begin{abstract}
We propose a novel system to help fact-checkers formulate search queries for known misinformation claims and effectively search across multiple social media platforms. 
We introduce an adaptable rewriting strategy, where editing actions for queries containing claims (e.g., swap a word with its synonym; change verb tense into present simple) are automatically learned through offline reinforcement learning. Our model uses a decision transformer to learn a sequence of editing actions that maximizes query retrieval metrics such as mean average precision. We conduct a series of experiments showing that our query rewriting system achieves a relative increase in the effectiveness of the queries of up to 42\%, while producing editing action sequences that are human interpretable. 
\end{abstract}

\maketitle

\section{Introduction}

With the wide spread of both human and automatically generated misinformation, there is an increasing need for tools that assist fact-checkers while retrieving relevant evidence to fact-check a claim. 
This process often involves searching for similar claims across social media using initial clues or keywords based on users' intuition. 
However, the available mechanisms for 
search on social media sites are
often platform-specific, with restrictions on the allowed number of search queries and access to retrieved documents. This can be attributed, among others, to the dynamic nature of social media feeds, the differences among users' interactions, and the architectural differences in how platforms perform search on their data. As a result, optimizing for arbitrary black-box search end-points containing ever-changing and different document sets means that a generic claim rewriter operating across all search end-points has a high chance of being sub-optimal.

\begin{figure}[t]
    \centering
    \includegraphics[width=0.48\textwidth]{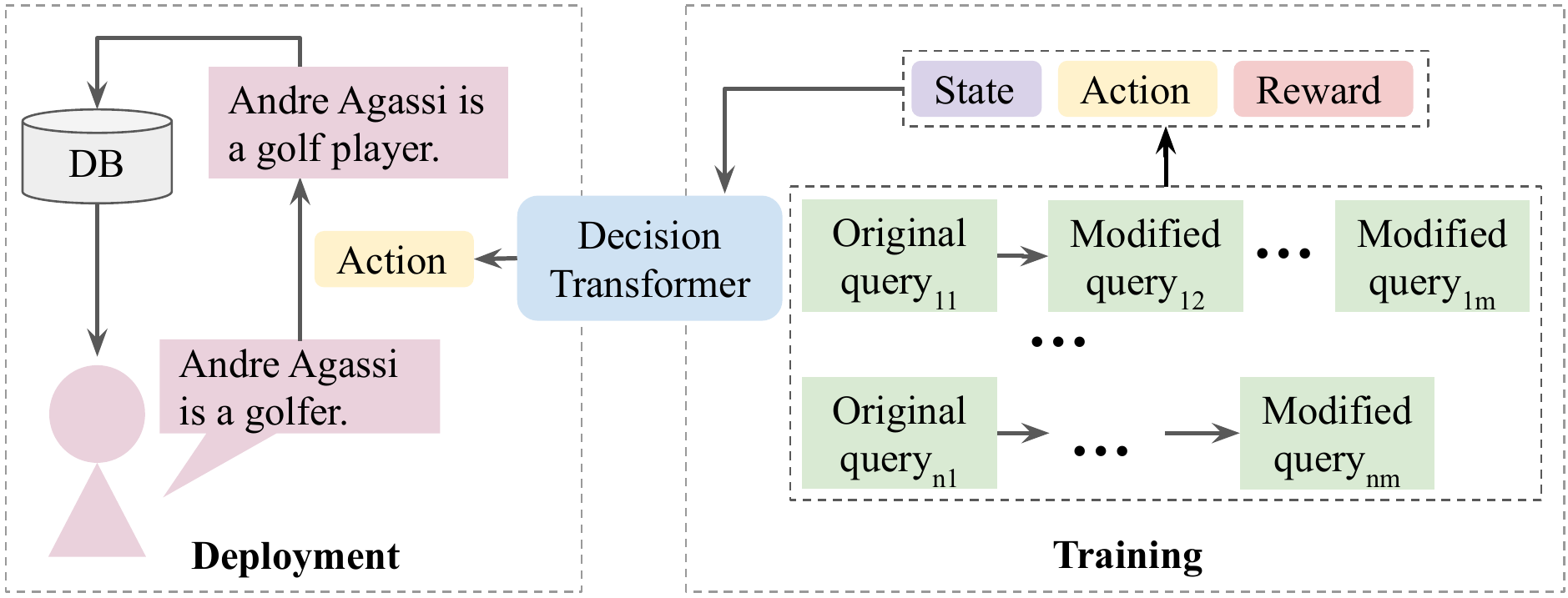}
    \caption{Overview of our proposed approach: we train a decision transformer with ``state'', ``action'' and ``reward'' sequences discovered by searching the space of potential query edits. During the deployment stage, the decision transformer predicts action(s) to rewrite the claim into a more effective query.}
    \label{fig:concept-diagram}
\end{figure}

To address these challenges, we draw upon a direct collaboration among fact-checkers and NLP researchers, and  introduce an adaptive claim rewriting system that can be used for effective misinformation discovery. We develop an interface in which users can edit individual tokens in the input claim using a predefined set of actions, and obtain updated queries leading to different levels of retrieval performance. Using this environment, we build a system that learns to rewrite input claims as effective queries by leveraging reinforcement learning (RL) to maximize desired retrieval metrics (e.g., average precision at K (AP@K)). An offline RL agent is then trained to learn the best editing sequences using a decision transformer model \cite{NEURIPS2021_7f489f64} as shown in \Cref{fig:concept-diagram}.

Given the limited access to social media search APIs, we use off-the-shelf retrievers such as BM25 \cite{robertson2009probabilistic} and approximate K-nearest neighbours (kNN) \cite{malkov2018efficient} to simulate platform search end-points. Our system is trained using a modified version of FEVER~\cite{thorne-etal-2018-fever}, a well known misinformation dataset containing a mix of true and false claims linked to Wikipedia evidence sentences. 
We transform FEVER claims into sequences of (claim, edit action, reward) triplets by using Breadth First Search (BFS) and heuristics such as constraining search space depth. 
These triplets are  used to train a decision transformer model to autoregressively predict a sequence of editing actions leading to retrieval improvements. 

Through several experiments, we show that our query rewriting approach leads to relative performance improvements of up to 42\% when compared to using the original claim. We also find that a simplified version of this approach--- i.e., fine-tuning a classifier to predict a single edit, leads to comparable performance while being more resource efficient during training and inference. We conduct ablation experiments to further evaluate the model performance across several settings, including variations on the retriever type, the reward metric, and the presence of negative training examples. 

To the best of our knowledge, our system is the first to leverage RL to learn to edit text from a set of human-readable actions only. From a practical perspective, it provides initial experimental evidence on the potential of interpretable systems in helping users, including fact-checkers, media writers, and platform trust and safety teams, to more effectively discover misinformation on the Internet. 

\section{Prior Work}

Our work is closely related to three previous research directions.
\paragraph{Finding Similar Claims.} 

The problem of finding similar claims has been explored from the perspective of system building, and supports a key step in human-led fact-checking \citep{nakov2021automated}. \citet{shaar-etal-2020-known} conducted retrieval and ranking of previously fact-checked claims given an input claim to detect debunked misinformation in English. \citet{kazemi-etal-2021-claim} tackled a similar problem in non-English languages. \citet{kazemi2022matching} investigated systems and models for finding applicable fact-checks for tweets.

While most prior work on this area has focused on building retrieval systems to identify similar claims, 
our work focuses on query rewriting to assist fact-checkers in the discovery of misinformation. During this process we assume that the retrieval system is a black-box to which we only have search access. 

\paragraph{Query Rewriting.} 

Query reformulation methods such as relevance feedback and local or global query expansion have been well-studied within the information retrieval literature. \citet{lavrenko2001relevance} proposed the \textit{relevance model}, an unsupervised local expansion method in which the probability of adding a term to the query is proportional to the probability of the term being generated from language models of the original query and the document the term appears in.  \citet{cao2008selecting} proposed a supervised pseudo relevance feedback in which expansion terms are selected by a classifier that determines their usefulness to the query performance. 
\citet{li2014req} introduced REC-REQ, an iterative  double-loop relevance feedback process in which a user provides relevance feedback to a classifier that is trained to identify relevant documents.

RL approaches have been previously applied to query rewriting. \citet{nogueira-cho-2017-task} and \citet{narasimhan-etal-2016-improving} used RL to learn to pick terms from pseudo-relevant documents that upon addition to the query improve retrieval performance metrics such as recall. In more recent work, \citet{wu2021conqrr} proposed CONQRR, a system that rewrites conversational queries into standalone questions. The authors first trained a T5 model to generate human rewritten queries for the QReCC dataset \cite{anantha-etal-2021-open} and then used them to generate candidate queries, which are selected based on maximizing search utility by an RL agent.

A key difference between our method and prior work is that we do not use information from the retrieved documents to reformulate queries as the queries themselves are the only input to the model. 

\paragraph{Text Editing Models.}\label{sec:text-editing-models}
Also related to our work is research done on ``text-editing'' models~\cite{malmi-etal-2022-text}. 
This line of research has gained traction in recent years as models such as EdiT5 and LEWIS~\cite{mallinson2022edit5,reid-zhong-2021-lewis} promise hallucination-free and controlled text generation for tasks where the input and output texts are similar enough so that a model can learn to transform the input into the output by applying a limited number of editing actions.  \citet{stahlberg-kumar-2020-seq2edits} proposed Seq2Edits, a  fast text-editing model for text generation tasks such as grammatical error correction and text simplification. Seq2Edits uses an edited transformer encoder and decoder to generate sequences of edits for the positions in the input text that need to be altered with suggested new tokens. 
\citet{reid-zhong-2021-lewis} introduced a multi-span text editing algorithm that uses Levenstein edit operations for the tasks of sentiment and politeness transfer in text, based on the intuition that text style transfer usually can be done with a few edits on the input text. Overall, text-editing models are usually faster than other sequence generation models such as seq2seq, since they only predict actions on a few input tokens rather than regenerating the whole sequence.

\section{Methods}

\subsection{Problem Definition}
In this paper, we focus on the task of query rewriting for discovering similar claims from an opaque search end-point. We have a collection of input claims ($C_1, C_2, ..., C_n$) that contain at least one fact-checkable claim. For any given claim $C_i$ in the collection, there exists one or more collections of similar claims ($SC_{i1}, SC_{i2}, ..., SC_{im}$), either supporting or refuting the claim in-part or as a whole.  
The RL agent operates on a fixed set of actions $\mathbf{A} = \{A_1, A_2, ..., A_k\}$ that can be applied to any of $C_i$'s tokens ($T_{i1}, T_{i2}, ..., T_{iq}$), where $k$ is the number of possible actions, $q$ is the number of tokens in $C_i$. We rewrite the query by applying the sequence of actions ($A_{ij}, 1\leq i\leq k, 1\leq j \leq q$) generated by the RL model to the original query. We can then use this improved query to retrieve related evidence statements.

\subsection{Model Overview}
\label{subsec: model-overview}
\begin{figure}[t]
    \centering
    \includegraphics[width=\linewidth]{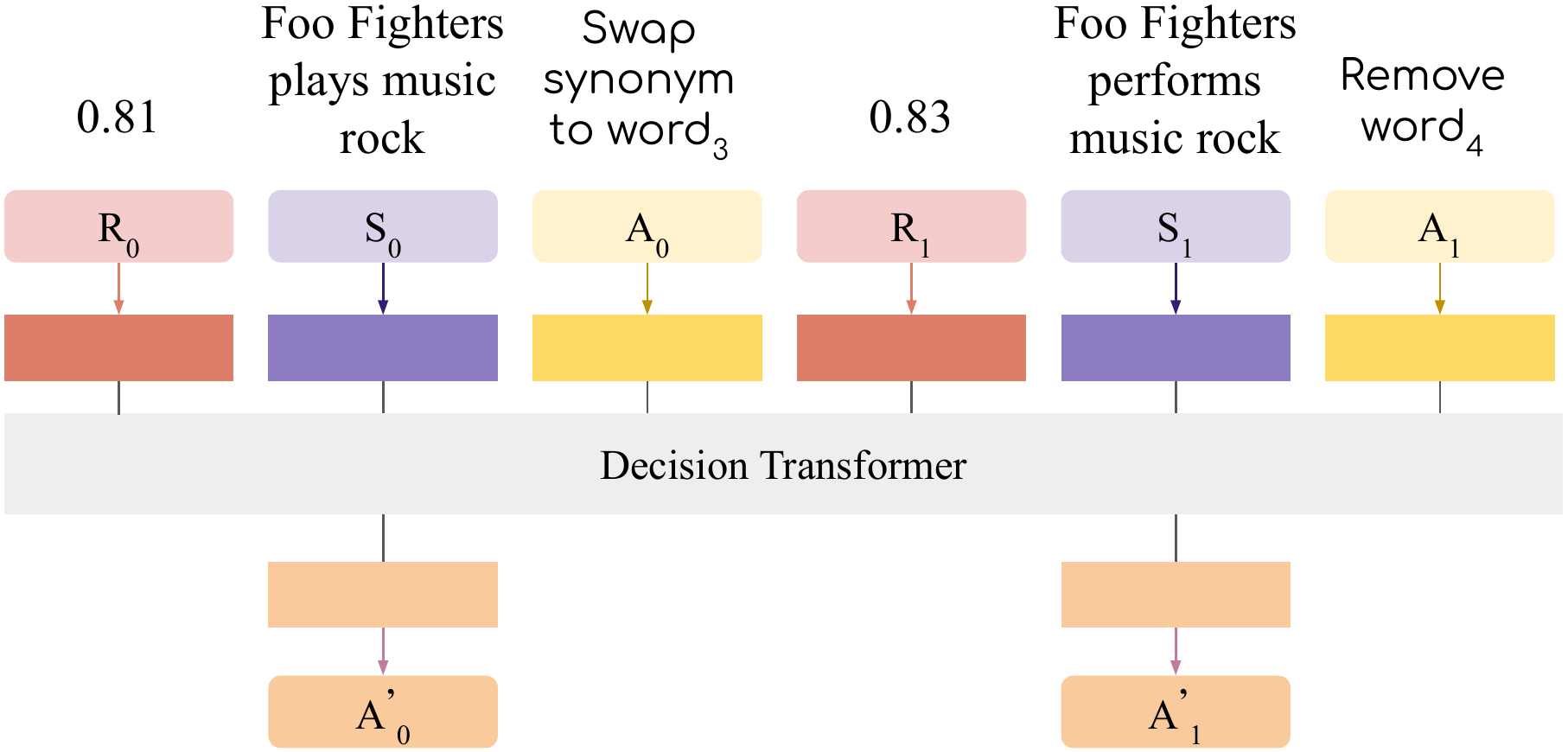}
    \caption{Model architecture. R, S, A represent reward, state and action, respectively. For instance, the state $S_0$ corresponds to a query, and the reward $R_0$ is the retrieval score such as AP@K. After we apply the action $A_0$ to the query $S_0$, the query becomes $S_1$. In inference time, the decision transformer predicts a series of actions $\{A_0', A_1', \cdots\}$ to apply to the original query.}
    \label{fig:overview}
\end{figure}

Our  system rewrites a query using concepts from RL and query expansion.  We pass the query into a pre-trained language model and then use the pooled representation from the final layer as the state representation. We use a decision transformer architecture, where states, actions, and rewards are provided to the model as a flattened sequence. The decision transformer uses a decoder-only GPT architecture \cite{radford2018improving} to learn the optimal policy during training time. During inference time, it autoregressively predicts actions for a given state.  An overview of our model architecture is shown in \Cref{fig:overview}. Below, we describe important elements of the model architecture related to the query rewriting process.

\noindent \textbf{Rewriting Actions.} Queries are rewritten using the following set of actions.

\noindent (1) \textit{Add synonym}: adds the synonym of a selected word to the query. Previous work by work by~\citet{riezler-liu-2010-query, Mandal2019QueryRU}, showed that rewriting queries with synonyms can improve query performance by potentially resolving ambiguous query terms.

\noindent (2) \textit{Swap with synonym}: replaces a specific word from the query with its synonym. This action has the same goal as \textit{add synonym}. Note that it includes the removal of the original token \textit{remove(original\_token)} .

\noindent (3) \textit{Change tense to present simple}: changes verb tense into present simple for selected verbs in the input. Changing verbs to their morphological variants has been previously found useful for query rewriting ~\cite{rafiei2009wild, haviv-etal-2021-bertese}.

\noindent (4) \textit{Remove}: deletes selected words from the query. Previous work has found that deleting words in queries can lead to higher  coverage of the search content~\cite{Jones2003QueryWD}.

We implement these actions using WordNet \cite{miller-1994-wordnet} and the spaCy's part-of-speech tagger. Note that only certain actions are permitted for each part of speech tag: verbs support all four actions, nouns, adjectives and adverbs support all actions except changing verb tense, and stop words and other parts of speech support only the remove action.

\label{sec:trainingoverview}

\noindent \textbf{State Representation.}  
We use sentence embeddings of the input claim as its state representation.
An input claim $C_i$ is passed through a Sentence-BERT \cite{reimers-gurevych-2019-sentence} network . The weights of the underlying pretrained language model (LM) are fine-tuned together with the decision transformer.

\noindent \textbf{Action Representation.} Our action space is two-dimensional: the first dimension represents the four action types (\textit{add synonym}, \textit{swap with synonym}, \textit{change tense to present simple} and \textit{remove}) and the second dimension represents the position of the token under edit, up to a maximum of 32 tokens. We pack these dimensions into a single dimension by taking their product, as shown in Table~\ref{tab:flat_actions}. Similar to the original implementation of the decision transformer, we pass the actions through a learned embedding layer to obtain an action vector representation. 


\begin{table}[t!]
\small
\centering
\begin{tabular}{c|c}
\toprule 
\toprule
Action \# & Edit, Position \\ \hline 
action 0:   & \textit{swap with synonym, position 0}  \\
action 1:   & \textit{swap with synonym, position 1}  \\
...        &         ...           \\
action 32:   & \textit{add synonym, position 0}  \\
action 33:   & \textit{add synonym, position 1}  \\
...        &         ...           \\
action 126: & \textit{remove, position 30} \\   
action 127: & \textit{remove, position 31} \\
\bottomrule
\bottomrule
\end{tabular}
\caption{A 2D space of actions types and token indices mapped onto a linear action space.}
\label{tab:flat_actions}
\end{table}

\noindent \textbf{Rewards.} We use the retrieval score for the edited query as the system reward at time step $t$.  Since the decision transformer uses returns-to-go to inform the model about future rewards, we use the sum of future rewards as a returns-to-go $R_t = \sum_{t'=t}^{T} r_{t'}$. We also experimented with a delayed reward strategy, where we set the returns-to-go for the last time step to be the maximum score for given claim seen during the data generation process, and zero for intermediate steps. During inference, we initialize returns-to-go to the maximum reward and decrease it by the achieved score after we apply an action.

\subsection{Retriever}\label{sec:retrieval}
Since access to social media API search endpoints is limited, it is difficult to train an RL agent on top of them. Furthermore, the changing nature of misinformation on social media is another important factor to take into account, given that misinformative posts are periodically removed from social media platforms and are thus no longer available once fact-checked. These issues made us opt for a simulated search environment, with the added benefit of making our methods adaptable to arbitrary search endpoints. We experiment with two main systems:  

\noindent \textbf{BM25}. A retriever frequently used in the literature as a retrieval baseline~\cite{robertson2009probabilistic}. We use the Elasticsearch implementation of BM25 with the default parameters.


\noindent \textbf{Approximate kNN.} 
A kNN retriever implemented using Elasticsearch's dense vector retrieval. We encode our data using pre-trained Sentence-BERT \cite{reimers-gurevych-2019-sentence} and use the embeddings to conduct an approximate kNN search using the Hierarchical Navigable Small Worlds (HNSW) algorithm \cite{malkov2018efficient}.

\section{Data}

\subsection{FEVER Dataset}

The FEVER dataset \cite{thorne-etal-2018-fever} is a collection of manually written claims from Wikipedia that are connected with evidence sentences that either ``support'' or ``refute'' them. Since we are interested in claims linked to related evidence, we discard the claims in the dataset labeled as ``NotEnoughInfo.'' This leaves us with 102,292 claims in the training and 13,089 claims in the development sets. \cite{schuster-etal-2019-towards} identified issues caused by the construction processes of the original FEVER dataset such as uses of negation in claims being heavily correlated with the ``refute'' outcome, therefore causing a ``claim only'' fact verification system to performs as well as an evidence-aware fact verification system. However, since our work is not concerned with the fact verification application of FEVER, we do not find this to be an issue.

FEVER is a well-known dataset among the misinformation and fact-checking communities. Even if FEVER is not  a social media dataset, it is nonetheless based on user-contributed data, and thus we believe that the findings obtained using this dataset can be generalized to claims on social media platforms with minor domain-specific revisions, especially since  the linguistic structure of claims and discussions around them is similar to the claims in the FEVER dataset.

\begin{figure}[t]
    \centering
    \includegraphics[width=\linewidth]{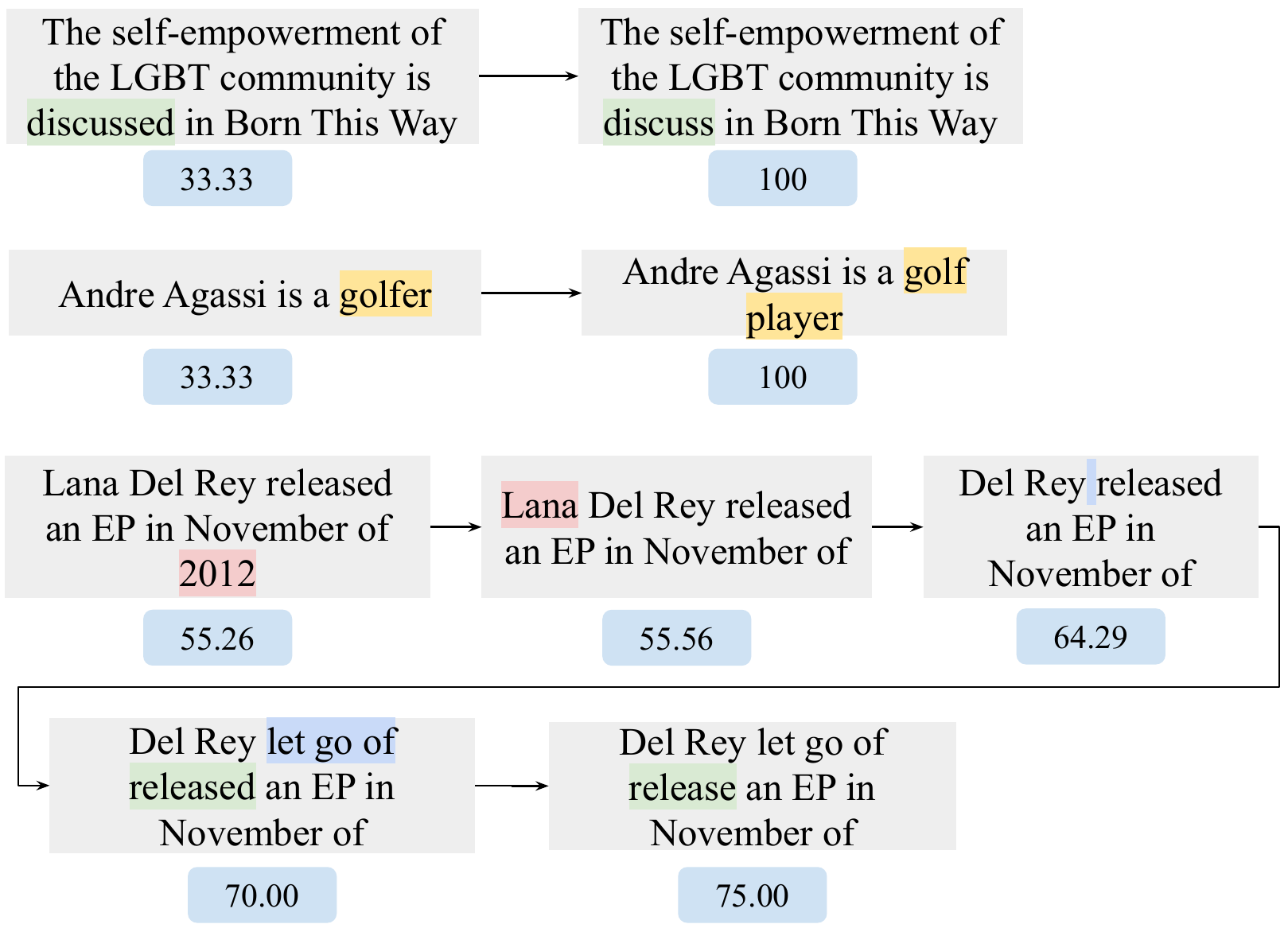}
    \caption{Sample sequence of claims generated by different actions: \remove, \present, \swap, \add~highlight \cremove{the token to remove}, \cpresent{the corresponding tokens to change tense} as well as \cswap{to swap to its synonym}, or \cadd{the corresponding places to add synonym} in red, green, yellow and blue, respectively. We report the corresponding AP@50 scores below each claim. \Cref{subsec: model-overview} \textbf{Rewriting Actions} provides intuitions of why these actions lead to better scores.}
    \label{fig:RL-data-generation-examples}
\end{figure}

\subsection{Generating RL-Friendly Training Data}
\label{subsec: generating-RL-friendly-training-data}

To generate training data, we transform FEVER pairs (claim, evidence set) into sequences of editing actions that improve upon the original query. These transformations are obtained by exploring the state space of possible outcomes after applying different permutations of edits on the initial claims. We use a Breadth-First Search (BFS) strategy that applies editing actions to an input claim $C_{i0}$ and finds the collection of the action sequences of $(C_{ij-1}, A_j, C_{ij}, R)$ that can improve the initial claim, where $C_{ij}$ is the generated claim after applying the edit $A_j$ to the claim $C_{ij-1}$, and $R$ is the reward of $Q(C_{ij})$ (querying retriever with $C_{ij}$).

Although understanding the effects of different search algorithms on our model remains an interesting problem for future work, our experiments show that using simple heuristics on BFS search is effective while generating training data from the FEVER dataset. For instance, we find that limiting the depth of the breadth-first exploration to K levels is effective for improving the query results. Also, when conducting parallel runs on different sections of the dataset, even for K = 4, the vanilla depth-limited BFS takes a half to 2 days to generate the training data. 
Additionally, we find that restricting the state-space search to include only improvement edits at every step reduces the size of the search space.
We also prune search paths leading to minor improvements (i.e. less than 3\%) or at random in 5\% of instances. Since most edits do not lead to significant improvements, it is unlikely we skip meaningful paths during the search. Finally, we only include sequences with the highest gains through serial edits, e.g. picking the top 50 or 100 most beneficial editing sequences for each claim, in our training set. Overall, these heuristics improve the generation speed and quality of the training instances.

Moreover, our ability to learn good editing actions depends on how well we can generate training examples. By setting $K$, the maximum depth for search to 4, we are able to get improvements up to 41.21 AP@50 scores for 45,658 claims, on average, against the BM25 retriever. We also discard training examples with reward values already at maximum, since it is impossible to improve beyond the perfect score, and also edited claims leading to no improvement. \Cref{fig:RL-data-generation-examples} shows examples of the sequences of claims generated by different actions. \Cref{tab:RL-data-statistics} reports the distribution of actions as well as the average improvement of AP@50 scores for each action when tested against the BM25 retriever. 

\begin{table}[t]
    \small
    \centering
    \begin{tabular}{crrrr}
        \toprule
        \toprule
         & \cremove{Remove} & \cswap{Swap\_Syn} & \cadd{Add\_Syn} & \cpresent{Present}  \\
         \midrule
      \% & \textbf{76.64} & 13.71 & 6.36 & 3.30 \\
      $\Delta$   & 11.56 & 12.36 & \textbf{12.84} & 12.28  \\
      \bottomrule
      \bottomrule
    \end{tabular}
    \caption{Percentage (\%) and mAP@50 ($\Delta$) improvements per rewriting action  
    against the BM25 retriever.}
    \label{tab:RL-data-statistics}
\end{table}


\section{Experiments}

We perform several experiments to determine the effectiveness of our adaptable query rewriting strategy. As a search environment, we use the BM25 and approximate kNN information retrieval methods described in Section \ref{sec:retrieval}.

\subsection{Experiment Settings}

During our experiments, we use the  original decision transformer implementation.\footnote{\url{https://github.com/kzl/decision-transformer}} We use a 6-layer decoder-only transformer with 8 heads, embedding dimension of 768. We set $K$ (also called a block size) to be the maximum number of edits to the original query. We pad all sequences  shorter than $K$. After flattening all the returns-to-go, states and actions, our sequence becomes of length $K*3$.

We use the \textit{all-mpnet-base-v2} embedding model from the Huggingface's sentence transformers library.\footnote{\url{https://huggingface.co/sentence-transformers}} We also experimented with the all-MiniLM-L12 model from the sentence transformers, but the results were  worse, possibly because of all-MiniLM-L12 being a smaller model. Our intermediate state representations for an input claim is a vector of size $768$. Our model is trained with cross entropy loss for 5 epochs performed on one Nvidia 2080Ti GPU. 

\subsection{Results}

\begin{table}[t]
    \centering
    \small
    \begin{tabular}{lc}
        \toprule
        \toprule
        Model &  mAP@50\\
        \midrule
        Original Claim & 26.83\\
        Random Baseline & 21.44\\
        Decision Transformer$_\text{sparse reward}$ & 32.43 \\
        Decision Transformer$_\text{dense reward}$ & \textbf{33.14} \\
        Fine-Tuned One Action Classifier & 31.95 \\
        \bottomrule
        \bottomrule
    \end{tabular}
    \caption{Experiment results with BM25 as retriever. 
    }
    \label{tab: model-results}
\end{table}
Results in Table \ref{tab: model-results} show that the decision transformer model with fine-tuned state embeddings and dense rewards outperforms  all systems with BM25 as retriever and AP@50 as reward. The same model trained with sparse rewards---hiding the intermediate rewards during training---does slightly worse than the dense reward setting, suggesting that providing more granular information about each action's reward during training brings performance advantages. Both models turn the input into a significantly more effective query with performance improvements of up to 23\% (relatively) as compared to just searching for the original claim. According to \Cref{tab:ablations} these gains are the highest for kNN as retriever and recall as reward. Table \ref{tab: model-results} also shows that performing a random sequence of edit actions negatively affects performance. This suggests that there is a ``query improvement process" that needs to be learned and applying a random sequence of edits by itself does not bring any inherent advantages, i.e. our systems do well not because there is an inherent gain in how we transform the problem, since if that was true, applying random action sequences should have yielded improvements over the claim baseline, which it did not.

\subsection{Analysis}

\begin{figure}[t]
    \centering
    \includegraphics[width=\linewidth]{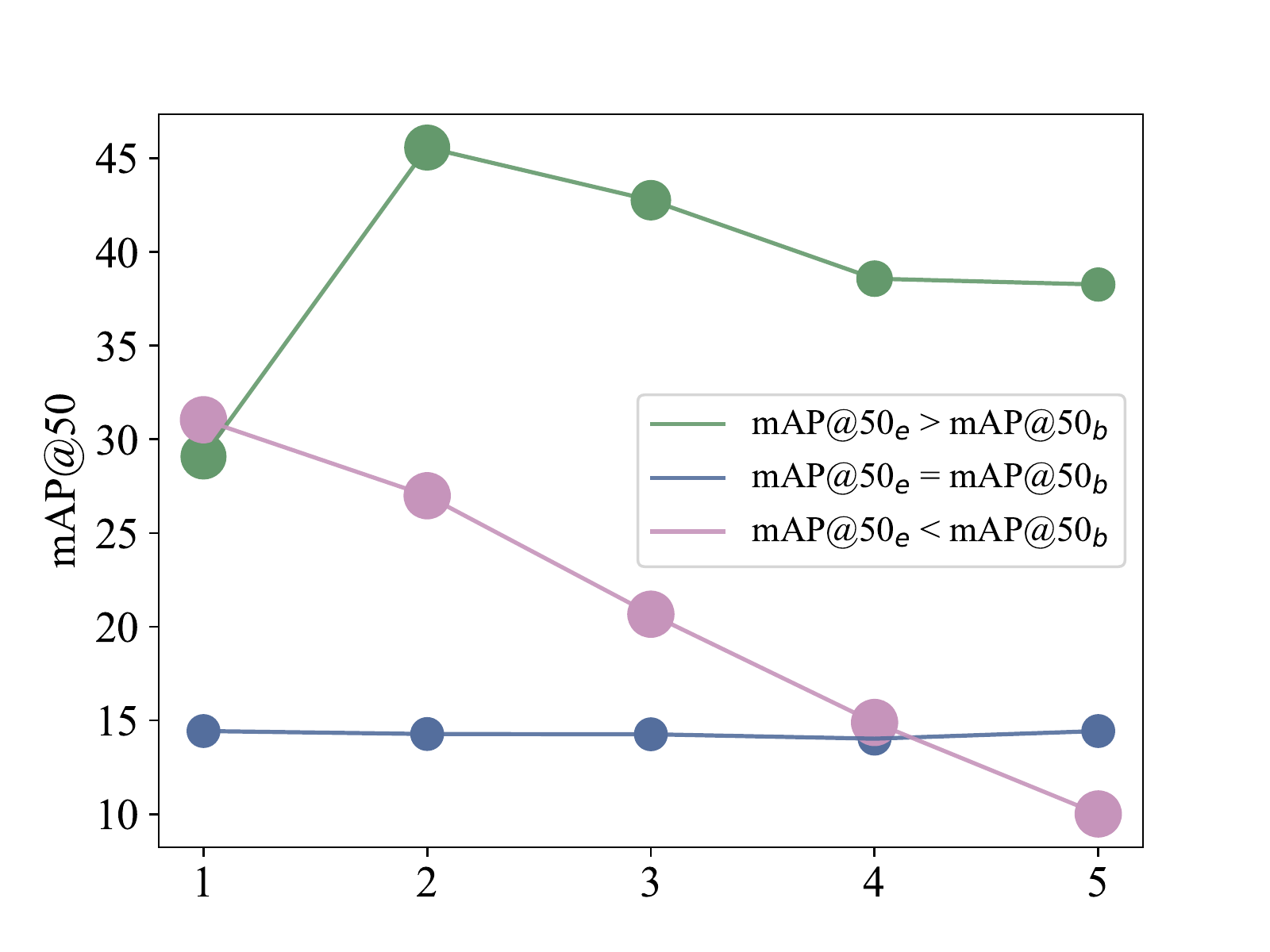}
    \caption{mAP@50 scores for all rewritten queries in the development set run against BM25. The x-axis indicates the claim rewriting sequence. 
    The size of each circle represents the number of queries at each turn.
    The subscripts ``e'' and ``b'' correspond to ``end'' and ``beginning'' of the claim rewriting sequence, respectively. 
    }
    \label{fig:sampled-step-map50-changes}
    \setlength{\belowcaptionskip}{-10pt}

\end{figure}

\Cref{fig:sampled-step-map50-changes} shows the mean AP@50 (mAP@50) score changes for all the generated sequences for the experiment of decision transformer with sparse reward. We plot the mAP@50 scores for queries generated at each step, where the x-axis shows the number of edits, with 1 representing the original claim and 5 the final rewritten query. The size of the circle indicates the number of queries at each turn. If the claim achieves the perfect score, no further rewrites will be generated in the next turn and we stop early. We observe sequences with {\bf improved mAP@50 scores} shrink along the turns. This indicates that some claims reach a perfect mAP@50 score after only one or two modifications. In contrast, for sequences with a {\bf decreased mAP@50 scores}, the circle sizes remain the same while performance drops. This suggests that for such claims, the more the model modifies it, the worse its performance is. For the sequence of claims with {\bf the same mAP@50 scores} at the beginning and the end, there is a slight up and down for the slopes for the lines in between. This suggests that there are some sequences where the modified query achieves a better score while later modifications hurt the performance or vice versa. However, such scenarios are rare. Of the 13,089 claims in the development set, 1541 claims have the same AP@50 scores at the beginning and the end. Among these, 1243 are constant along the entire sequence and 271 have minor score changes, as reflected in \Cref{fig:sampled-step-map50-changes} as the blue line (mAP@50$_\text{\textit{e}}$ = mAP@50$_\text{\textit{b}}$).

\begin{figure}[t]
    \centering
    \includegraphics[width=\linewidth]{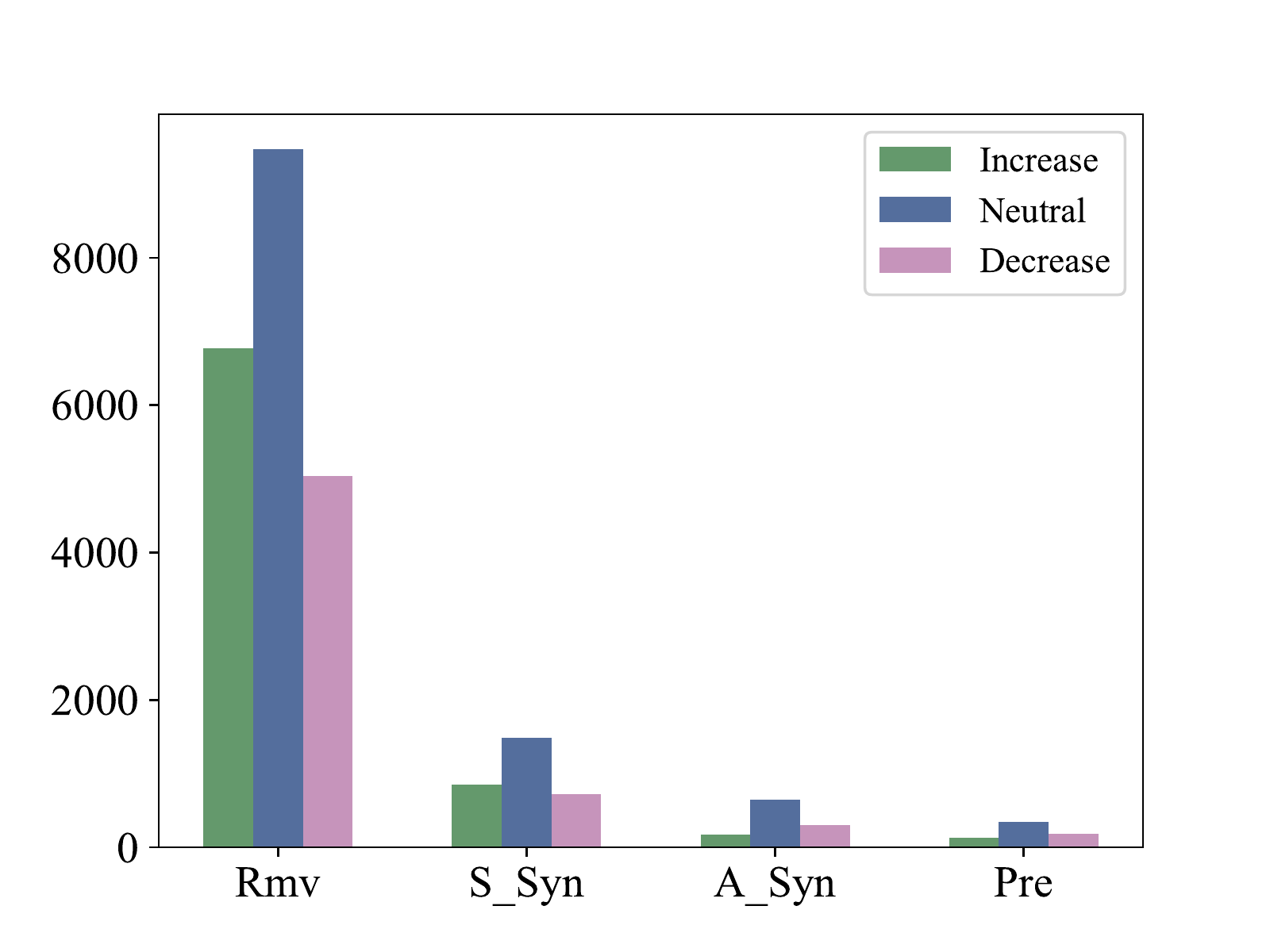}
    \caption{Distribution of predicted actions (\textit{remove}, \textit{swap with synonym}, \textit{add synonym}~and \textit{change to present tense}) with AP@50 reward and BM25 retriever.}
    \label{fig:action-distribution-generated-file}
\end{figure}

\begin{figure}[t]
    \centering
    \includegraphics[width=\linewidth]{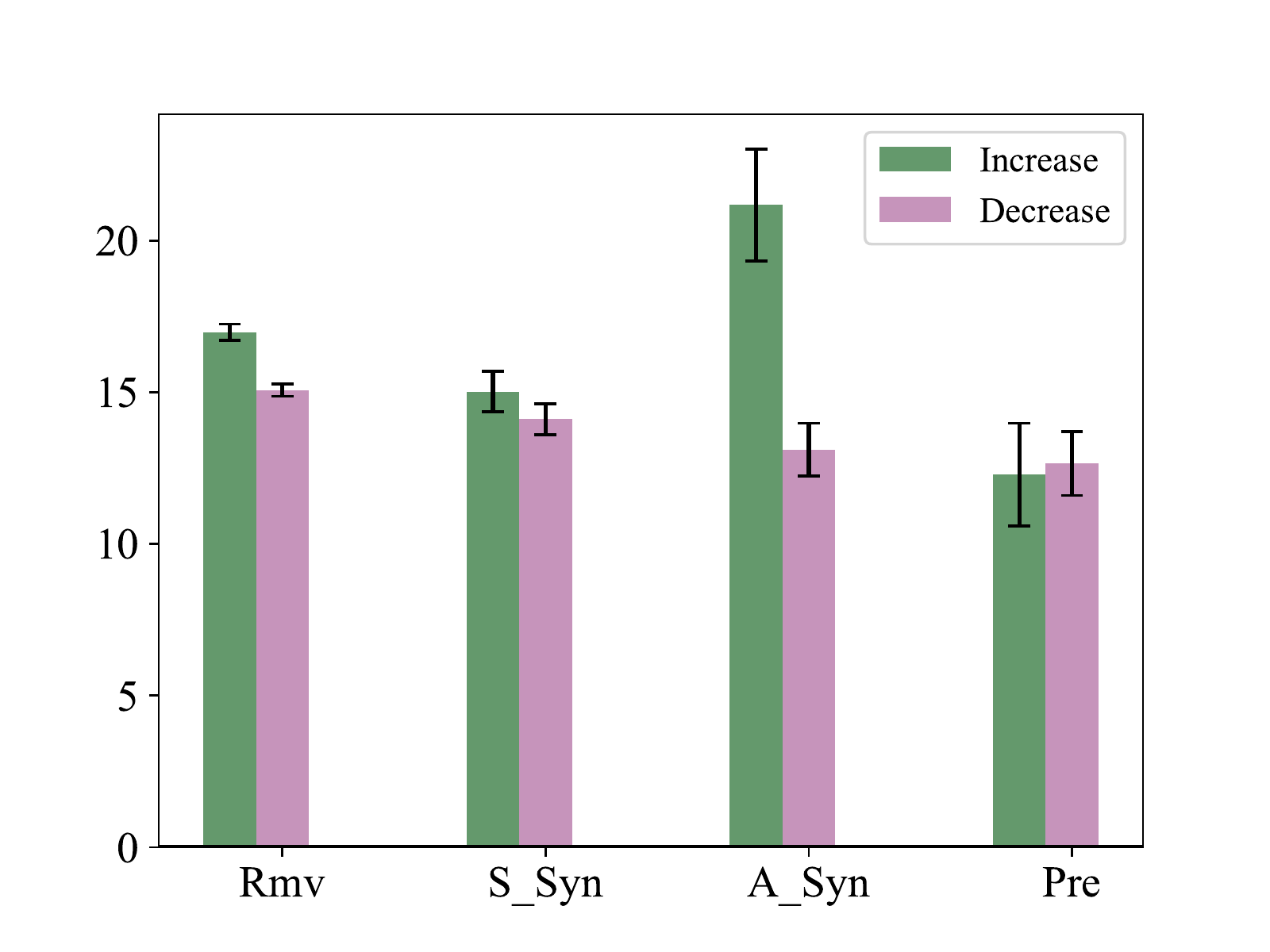}
    \caption{Average change in AP@50 scores of the predicted actions (\textit{remove} (Rmv), \textit{swap with synonym} (S{\_}
Syn), \textit{add synonym} (A{\_}
Syn) and \textit{change tense to present simple} (Pre)) against BM25.  Statistics for actions with no changes in AP@50 are excluded as this results in 0 scores.}
    \label{fig:avg-score-changes-generated-file}
\end{figure}

\Cref{fig:action-distribution-generated-file} shows the distribution of actions in the model output corresponding to increased performance, no changes in performance, and decreased performance. We can see that most of the actions lead to no performance change. The \textit{remove} and \textit{swap with synonym} actions result more often in increases in performance than decreases. In contrast, \textit{add synonym} and \textit{change tense to present simple} more often result in performance reduction. \Cref{fig:avg-score-changes-generated-file} shows the average change per action. In this plot we observe that the net performance changes for \textit{remove} and \textit{swap with synonym} are positive, with an average of 1.83 and 0.88, respectively. The net performance changes for \textit{add synonym} and \textit{change tense to present simple} are negative, with an average of -0.34 and -0.98, respectively. We hypothesize that the model does not learn \textit{add synonym} and \textit{change tense to present simple} actions well due to the sparsity of such examples in the data as shown in \Cref{tab:RL-data-statistics}. We further discuss the importance of these actions in  \Cref{sec: discussion}. 

\begin{table*}[t]
\centering
\small
\begin{tabular}{c | c  c  c | c  c  c}
\hline 
\hline 
\multirow{ 2}{*}{Retriever(Query)} & \multicolumn{3}{c|}{↑ rewards only} & \multicolumn{3}{c}{↑ + ↓ rewards} \\
 & mAP@50            & Recall         & RR             & mAP              & Recall           & RR              \\
\hline
BM25(\textbf{RL}[Claim])           & \textbf{32.43} & \textbf{35.8}  & \textbf{30.23} & 31.50             & 32.82            & 29.80            \\
BM25(Claim)               & 26.82          & 29.68          & 22.30           & 26.82            & 29.68            & 22.30        \\
kNN(\textbf{RL}[Claim])            & \textbf{36.69} & \textbf{36.95} & 29.17 & 34.49            & 35.06            & \textbf{29.79}           \\
kNN(Claim)                & 28.40           & 25.93          & 21.27          & 28.40             & 25.93            & 21.27 \\
\hline 
\hline 
\end{tabular}
\caption{Ablation experiments, RR refers to reciprocal rank.}
\label{tab:ablations}
\end{table*}

\subsection{Ablations}

We conduct ablation experiments to evaluate the ability of our system in adapting to arbitrary endpoints and different performance metrics. Although the space of possible ablations is far larger than what we present here, we pick three dimensions of ablations that could be useful for practitioners and future researchers: (i) retriever type (BM25 or kNN), (ii) reward metric (average precision, recall, reciprocal rank) and (iii) presence of negative training examples.  

Table~\ref{tab:ablations} shows the results on each ablation when compared against a baseline of just using the initial claim. Across different metrics and retrievers we observe improvements in query performance: our system improves the original claim of up to 11\% absolute recall points (42\% relative improvement) and works on both BM25 and kNN retrievers. We also observe that the inclusion of  training sequences with query performance decrease (negative training examples), consistently leads to performance decreases on all metrics and retrievers as compared to just training on positive edit sequences --with the only exception of querying kNN with RR as reward. We posit that this performance gap is due to the difference in data quality, i.e, providing our models with noiseless training signals leads to more effective queries. However, even in cases where we include negative training examples our models still meaningfully improve over the original claim.

\section{Discussion}
\label{sec: discussion}

\paragraph{Do we need to use (offline) RL for claim rewriting?} It can be argued that a computationally expensive RL agent for query rewriting could be replaced by more economic design choices such as a sequence labeling model by fine-tuning a pretrained language model. In fact, as we discussed in the prior work section (\ref{sec:text-editing-models}), researchers have indeed taken several different approaches for training neural text-editing models. In order to dig deeper into this question, we chose AP@50 as reward and trained a classifier on only the first edit in the training instances as 128-way classification (4 actions * 32 tokens), and the resulting classifier performed slightly worse than the RL agent trained on the whole edit sequence. However, we also observe from Figure~\ref{fig:sampled-step-map50-changes} that when using the BM25 retriever and AP@50 as reward, the first action in training data is four times more effective than the following three actions on average, which means that the comparison between the classifier and the RL agent might not be a fair one. However, we also interpret the strong performance of the classifier as a more \textit{efficient} alternative to training expensive reinforcement learning models. We leave a deeper comparison of the capabilities of sequence classification modeling and offline reinforcement learning for future work.

\paragraph{Are pretrained sentence embeddings good candidates for state representation?} In our initial set of experiments we used frozen pretrained Sentence-BERT embeddings as state representation, and we did not see significant improvements over the initial claim. We observed a significant performance jump (~5 mAP@50 points) once the sentence embeddings were also trained alongside the RL agent. This improvement highlights {\bf the importance of state representation and shows that task-specific embeddings perform better than general-purpose  embeddings}. This finding also indicates that the presence of Wikipedia data in the training data of LLMs does not simplify our task. Furthermore, there is significant prior work emphasizing the role and difficulty of the combinatorial and compositional nature of the state space in language tasks for reinforcement learning. \cite{cote2018textworld}, which also makes text-based RL agents a good choice for advancing our understanding of natural language.
\paragraph{What is the relation between query rewriting with sequence action learning and keyword extraction?} We find that some of our models predict the \textit{remove} action the vast majority of times, upwards of 80\% in the case of using BM25 as retriever and AP@50 as reward. This brings up a natural question around how our method compares with keyword extraction methods, since the prevalence of remove edits during inference suggests that our approach works similar to keyword extraction. Our initial experiments with KeyBERT \cite{grootendorst2020keybert} show that this is not the case as {\bf keyword extraction does not perform comparably with the claim baseline on BM25 and AP@50 as reward}. Although further analysis is required to make firm conclusions, it could be implied that including actions other than \textit{remove} for rewriting queries can bring in significant gains.

\section{Conclusion}
In this paper, we presented our findings on using an offline RL agent that learns editing strategies for query rewriting, so that fact-checkers can discover misinformation across social media platforms more effectively. 
Using a decision transformer, we showed that we can learn to rewrite misinformation claims by applying a series of  interpretable actions such as adding synonyms or removing specific words. These actions can transform the claims into more effective queries, leading to a relative performance increase of up to 42\%  over a simpler kNN retriever baseline. 
Additionally, we conducted further analyses and ablation studies to develop a better understanding of our system, which showed that its adaptable to a variety of metrics and search engines. Our findings are an initial step towards building AI-assisted technologies to help fact-checkers discover online misinformation more effectively.

\paragraph{Future Work.} While our work lays the grounds on using RL for building effective misinformation discovery tools, the practical application of our model requires further work to account for the limited access to social network APIs. This means additional constraints such as: (1) learning to rewrite claims under a fixed budget of training queries, and (2) learning without supervision. While there are already several solutions available for (2) \cite{shaar-etal-2020-known, kazemi-etal-2021-claim}, we believe (1) is an exciting area for further exploration. Additionally, we posit our approach to be applicable on languages other than English since the RL agent we train is mainly language-agnostic.

\section{Limitations}

Although we conduct ablations across several experimental settings, there are still important design decisions that require further research such as the design of action space and the utility of human-readable edits for explainability. Our action space is one choice among the set of many possible text editing actions, thus there could be more expressive or efficient action spaces that lead to more efficient queries. Although there is no need for the rewrites to be explainable, our method has the potential to be explainable since the rewriting process is entirely human-readable. To understand the explainability potential, a study augmented with human evaluation of the rewritten claims is necessary, which we leave for future work.

\section*{Acknowledgements}
We thank Lan Zhang, Qinyue Tan, and Davis Liang for their help and feedback to this project. This work was partially supported by a grant from Meta and an award from the Robert Wood Johnson Foundation (\#80345). Any opinions, findings, and conclusions or recommendations expressed in this material are those of the authors and do not necessarily reflect the views of Meta or the Robert Wood Johnson Foundation.

\bibliographystyle{ACM-Reference-Format}
\bibliography{anthology,custom}

\end{document}